\documentclass{article}
\usepackage[sorting=none]{biblatex}
\usepackage{algorithm}
\usepackage{algpseudocode}
\usepackage{PRIMEarxiv}
\usepackage{amsmath}
\usepackage[utf8]{inputenc} 
\usepackage[T1]{fontenc}    
\usepackage{hyperref}       
\usepackage{url}            
\usepackage{booktabs}       
\usepackage{amsfonts}       
\usepackage{nicefrac}       
\usepackage{microtype}      
\usepackage{lipsum}
\usepackage{fancyhdr}       
\usepackage{graphicx}       
\graphicspath{{media/}}     

\renewbibmacro*{name:andothers}{
  \ifboolexpr{
    test {\ifnumequal{\value{listcount}}{\value{liststop}}}
    and
    test \ifmorenames
  }
    {\ifnumgreater{\value{liststop}}{1}
       {\finalandcomma}
       {}%
     \andothersdelim\bibstring[\emph]{andothers}}
    {}}

\title{Fair and skill-diverse student group formation via constrained $k$-way graph partitioning}

\author{
  Alexander Jenkins$^1$, Imad Jaimoukha$^1$, Ljubisa Stankovic$^2$, Danilo Mandic$^1$ \\\\
  $^1$Department of Electrical and Electronic Engineering, Imperial College London, London SW7 2AZ, UK \\
   $^2$Faculty of Electrical Engineering, University of Montenegro, Podgorica, 81000, Montenegro \\
   E-mails: \texttt{\{a.jenkins21, i.jaimouka, d.mandic\}@imperial.ac.uk}, \texttt{ljubisa@ucg.ac.me}\\
}

\begin{document}
\maketitle

\begin{abstract}
Forming the right combination of students in a group promises to enable a powerful and effective environment for learning and collaboration. However, defining a group of students is a complex task which has to satisfy multiple constraints. This work introduces an unsupervised algorithm for fair and skill-diverse student group formation. This is achieved by taking account of student course marks and sensitive attributes provided by the education office. The skill sets of students are determined using unsupervised dimensionality reduction of course mark data via the Laplacian eigenmap. The problem is formulated as a constrained graph partitioning problem, whereby the diversity of skill sets in each group are maximised, group sizes are upper and lower bounded according to available resources, and `balance' of a sensitive attribute is lower bounded to enforce fairness in group formation. This optimisation problem is solved using integer programming and its effectiveness is demonstrated on a dataset of student course marks from Imperial College London. 
\end{abstract}


\section{Introduction}
Modern education often requires to form groups of students, for example, for study groups, tutorials or group projects. However, the formation of sub-groups of students is a manual, subjective and laborious task. The complexity of such a task is further increased by the necessity for group formation to satisfy constraints on group sizes and fairness requirements with respect to sensitive attributes such as gender. To this end, students are often allocated at random to a group or are allowed to select their own group.

Whilst defining the right group of students is subject to interpretation, it is widely agreed that diversifying skill sets within groups can create a stimulating and productive environment \cite{Jaques2007, katzenbach2015wisdom}. \textcite{danilopart1} introduced the idea of using unsupervised machine learning to identify student affinities from course mark data. Using a simulated dataset of $N$ students and their marks in $L$ courses, the authors considered every student to be a vertex in a graph, $G(V,E)$, where $V$ is a set of vertices connected by a set of edges $E$. Weighted edges connect pairs of students with the similarity between course marks encoded as the weight value. A Laplacian eigenmap \cite{6789755} was then used to reduce the dimensionality of the problem from $L$ to $M$, where $M \ll L$. By visualising the students in the reduced $M$ dimensional basis, students were found to cluster into their assigned affinity.

This work extends upon \cite{danilopart1} to introduce an unsupervised algorithm for fair and skill-diverse student group formation. This is achieved based on student course marks and sensitive attributes provided by the education office. More specifically, we use unsupervised dimensionality reduction as in \cite{danilopart1} to identify student affinities from data. The fair and skill-diverse group formation problem is then formulated as a constrained graph partitioning problem that can be solved using integer programming, whereby:
\begin{enumerate}
    \item Skill-diverse groups are found by maximising the distances between students' feature vectors in the Laplacian eigenmap;
    \item Fair groups are found by constraining the `balance' of sensitive attributes in the group relative to the population;
    \item Group sizes are constrained with upper and lower bounds.
\end{enumerate}

The remainder of the paper is organised as follows. In Section \ref{sec:background} the background information required to understand our algorithm will be discussed. In Section \ref{sec:methodology} the algorithm will be formulated. In Section \ref{sec:results} the algorithm will be tested on a dataset of student course marks from Imperial College London. 

\section{Background}\label{sec:background}
\subsection{Dimensionality reduction using graph Laplacian}
Dimensionality reduction refers to the transformation of high-dimensional data to a low-dimensional space such that useful information present in the data is preserved as much as possible. The transformation can be linear, such as the principal component analysis \cite{Jolliffe2014}, or non-linear, such as auto-encoders and Laplacian eigenmaps \cite{6789755}. The latter methods are referred to as `manifold learning' as they model the data as residing on a low-dimensional manifold embedded in a high-dimensional space. The Laplacian eigenmap is a dimensionality reduction method that discretely approximates the low-dimensional manifold by connecting data points in local neighbourhoods using a graph structure. It is chosen in this work due to its optimal locality-preserving property, which states that the data points which are close in the original $L$ dimensional space are also close in the reduced $M$ dimensional space.

For $N$ data points residing in an $L$ dimensional space, the position of the $m$-th data point given by the vector $\mathbf{r}_m \in \mathbb{R}^L$. A Laplacian eigenmap considers each data point as a vertex in a graph. An edge, $W_{mn}$, connects two vertices, $m$ and $n$, with a weight derived from the similarity between their vectors $\mathbf{r}_m$ and $\mathbf{r}_n$, such that vertices which are close in the high-dimensional space receive a large edge weight. A weighted adjacency matrix, $\mathbf{W} \in \mathbb{R}^{N\times N}$, is defined with elements $W_{mn}$, and contains the connectivity information for the graph. The graph Laplacian is defined as $\mathbf{L} = \mathbf{D}-\mathbf{W}$, where $\mathbf{D}\in \mathbb{R}^{N\times N}$ is a diagonal matrix with elements $D_{mm} = \sum_{n=1}^NW_{mn}$ representing the degree of each vertex. An eigen-decomposition of the graph Laplacian, $\mathbf{L} = \mathbf{U}\mathbf{\Lambda} \mathbf{U}^T$, yields the matrix of eigenvectors $\mathbf{U} \in \mathbb{R}^{N\times N}$ and the diagonal matrix of eigenvalues $\mathbf{\Lambda} \in \mathbb{R}^{N\times N}$ that are ordered in a decreasing manner. The Laplacian eigenmap represents each data point in a new $M$-dimensional space, where $M < L$, with a new basis for the $m$-th data point given by the \textit{spectral vector},
\begin{equation}
\label{eqn:spectral_vector}
\mathbf{q}_m = [u_1(m), ..., u_{M-1}(m)],
\end{equation}where the first smoothest eigenvector $\mathbf{u}_0$ has been removed \cite{6789755}.

\subsection{K-way graph partitioning}
Graph partitioning is a method for clustering vertices of a graph. For a graph, $G(V,E)$, a $k$-way partition is defined as the division of graph vertices into $k$ disjoint subsets $V^{(1)}, V^{(2)}, ..., V^{(k)}\subseteq V$ such that $V^{(i)} \cap V^{(j)} = \emptyset$  for all $i \neq j$ and $\bigcup_{\forall i} V^{(i)}=V$. An example of a graph partition is the minimum cut \cite{Goldschmidt1994}, which is defined as the minimum sum of edge weights that can be removed to divide graph vertices into $k$ disjoint subsets. The optimisation objective for graph partitioning can be designed / constrained to give desirable features of subsets. For example, \textcite{Labb2010} upper and lower bounded the size of vertex subsets in an integer programming framework. 

\subsection{Fairness metrics for clustering}
\textcite{NIPS2017_978fce5b} introduced the concept of `Balance' of a sensitive attribute, where a sensitive attribute must have approximately equal representation across all clusters. \textcite{BeraBalance} extended this work to introduce balance as a constraint for each cluster that can be upper and lower bounded. Balance of sensitive attribute $s$ in a group $c$ is defined as
\begin{equation}
\label{eqn:balance}
    B_{cs} = \min \left\{ R_{cs}, \frac{1}{R_{cs}} \right\},
\end{equation}
where $R_{cs} = \frac{a_{cs}}{a_s}$, $a_{cs}$ is the ratio of sensitive attribute $s$ in the group $c$, and $a_{s}$ is the ratio of sensitive attribute $s$ in the population.

\section{Methodology}
\label{sec:methodology}

\subsection{Dataset}
Anonymised datasets of student course marks from the Electronic and Electrical Engineering department at Imperial College London, United Kingdom, were analysed. The dataset corresponds to the first two years of course marks from the Electronic and Information Engineering (EIE) undergraduate degree stream. The EIE dataset consists of $N = 54$ students who sat $L = 23$ courses. Course marks in both datasets are given from $0-100$\%.

\subsection{Laplacian eigenmap construction}
A graph, $G_1$, was created whereby each student was defined as a vertex. The edge weights between each pair of students, $m$ and $n$, were defined as

\begin{equation}
\label{eqn:graph_construction}
W_{mn} = \left\{
  \begin{array}{ c l }
      \exp{ (\frac{Corr(\mathbf{r}_m, \mathbf{r}_n)^2}{A})} & \textrm{if} \quad Corr(\mathbf{r}_m, \mathbf{r}_n) \geq B, \\
      0 & \textrm{otherwise,}
  \end{array}
\right.
\end{equation}
where $\mathbf{r}_m$ and $\mathbf{r}_n$ are the vectors of course marks for students $m$ and $n$, with $r_m(k)$ denoting the mark of the $m$-th student in the $k$-th course; $Corr(\mathbf{r}_m, \mathbf{r}_n)$ is the correlation between the two vectors; and $A$ and $B$ are constants that influence the approximation of the underlying low-dimensional data manifold. In this work $A=10$ and $B=0.5$ were used. A weighted adjacency matrix for graph $G_1$ is denoted by $\mathbf{W} \in \mathbb{R}^{N\times N}$, with elements $W_{mn}$ computed from (\ref{eqn:graph_construction}).

A Laplacian eigenmap \cite{6789755} was constructed from the eigen-decomposition of the normalised graph Laplacian, $\mathbf{L}_{norm} = \mathbf{U}\mathbf{\Lambda} \mathbf{U}^T$, where $\mathbf{L}_{norm} = \mathbf{D}^{-\frac{1}{2}} \mathbf{L} \mathbf{D}^{-\frac{1}{2}}$ and $\mathbf{L}=\mathbf{D}-\mathbf{W}$. The $m$-th student was represented in a new reduced $M$-dimensional space, where $M < L$, with a new basis for this space given by the spectral vector $\mathbf{q}_m$ in (\ref{eqn:spectral_vector}) using the eigenvectors of the normalised graph Laplacian. For ease of visualisation and to preserve adequate information in dimensionality reduction, $M=3$ was chosen.

\subsection{Fair and skill-diverse group formation}
\paragraph{Definition of diversification.}
The Laplacian eigenmap is locally invariant, so students with similar course affinities (similar $\mathbf{r}_m$) will be closer together in the eigenmap (similar $\mathbf{q}_m$). This fact is exploited to construct skill-diverse groups of students. We define a group as diversified if it contains students with different course affinities. To measure the dissimilarity between two students, $m$ and $n$, the Euclidean distance in the Laplacian eigenmap space is used. This is given by
\begin{equation}
\label{eqn:dissimilarity_distance}
d_{mn} = \| \mathbf{q}_{m} - \mathbf{q}_n \|_2.
\end{equation}
A large distance will indicate a pair of students with different course affinities. This distance was used to construct edge weights in a new fully connected graph, $G_2$, where the vertices in $V$ are students.


\paragraph{Definition of fairness.}
Fairness was quantified based on the balance of each sensitive attribute, computed using (\ref{eqn:balance}) \cite{BeraBalance}. We define a group as fair if the ratio of the sensitive attribute $s$ in the group $c$ is equal to the ratio of the sensitive attribute in the population, i.e. $B_{cs} = 1$. In practice, a lower bound for balance is used to promote fairness, as it may not always be possible to achieve $B_{cs}=1$.

\paragraph{Optimisation problem.}
The fair and skill-diverse student group formation problem is formulated through the graph partition vector, $\mathbf{w} \in \{0, 1\}^{|E|}$, that maximises
\begin{equation}
\label{eqn:optimise}
\sum_{\{m,n\}\in E} d_{mn} w_{mn},
\end{equation}
where $E$ is the edge set of $G_2$, $w_{mn}$ is the element of the partition vector for the edge connecting the vertices $m$ and $n$, and $d_{mn}$ is the distance calculated in (\ref{eqn:dissimilarity_distance}). It is often desirable to constrain the resulting group sizes (e.g. due to limited resources within departments) and fairness of group formation with regard to a sensitive attribute (e.g. gender). The group sizes are constrained by an upper bound, $F_U \in \mathbb{Z}^+| 1 \leq F_U \leq N$, and a lower bound, $F_L\in \mathbb{Z}^+| 1 \leq F_L \leq N$, where $F_L \leq F_U$. Balance of a sensitive attribute, $s\in S$, within groups is constrained by a lower bound, $B_{L_s}$, where $B_{L_s} \in \mathbb{R}  |  0 \leq B_{L_s} \leq 1$ and $S$ is the set of all sensitive attributes to be considered.


We formulate the fair and skill-diverse student group formation problem as a constrained integer programming problem as in \cite{Labb2010}, given by
\begin{equation*}
\max_{\{w_{mn}\}_{\forall \{m, n\} \in E}} \quad \sum_{\{m, n\}\in E} d_{mn} w_{mn}
\end{equation*}
subject to
\begin{subequations}
\begin{align}
& w_{mn} + w_{mo} - w_{no} \leq 1 \quad \forall m,n,o\in V: m \neq n \neq o \label{eqn:constraint-triangle}\\
  & |\mathbf{w}(\delta(m))| + 1 \geq F_L \quad \forall m \in V  \label{eqn:constraint-size-lower}  \\
  & |\mathbf{w}(\delta(m))| + 1 \leq F_U \quad \forall m \in V  \label{eqn:constraint-size-upper} \\
  & w_{mn} \in \{0, 1\} \quad \forall{\{m,n\}}\in E \label{eqn:constraint-integer} \\
  & B_{L_s} \leq B_{cs}(m) \quad \forall s \in S, \forall m \in V \label{eqn:constraint-balance}.
\end{align}
\end{subequations}

The constraints (\ref{eqn:constraint-triangle}) are called triangle inequalities, which state that if the edge between vertices $m$ and $n$ is in a given partition, and the edge between vertices $m$ and $o$ is in the partition, then the edge between vertices $n$ and $o$ must be in the partition \cite{Labb2010}. The constraints in (\ref{eqn:constraint-size-lower}) and (\ref{eqn:constraint-size-upper}) correspond to the lower and upper bounds on the group sizes, respectively. The edges adjacent to vertex $m$ are given by $\delta(m) = \{\{m, n\} \in E|m \in V, n \in V-\{m\}\}\}$, and $\mathbf{w}(\delta(m))$ represents the subset of the partition vector with elements in $\delta(m)$. Therefore, $|\mathbf{w}(\delta(m))|$ is a count of the number of vertices connected to $m$. The constraints in (\ref{eqn:constraint-integer}) force the partition vector to have integer values. The constraints in (\ref{eqn:constraint-balance}) are our fairness constraints for graph partitioning. More specifically, this will lower bound the balance, $B_{cs}(m)$, of sensitive attribute $s$ for a group $c$ which contains vertex $m$. The balance, $B_{cs}(m)$, is computed as follows. Let $\mathbf{A}_s \in \{0, 1\}^{|V|}$ be the binary vector of vertex (student) attributes, which has value 1 if sensitive attribute $s$ is present. Let $\mathbf{A}_s(\delta (m))\in \{0,1\}^{|\delta(m)|}$ represent the subset of $\mathbf{A}_s$ for vertices connected to vertex $m$ by an edge, and $\mathbf{A}_s(m)$ designate the sensitive attribute of vertex $m$. The ratio of sensitive attribute $s$ in the group $c$ containing vertex $m$ is calculated as
\begin{equation}
\label{eqn:balance-constraint-calculation}
a_{cs} = \frac{\mathbf{A}_s(\delta (m)) \cdot \mathbf{w}(\delta(m)) + \mathbf{A}_s(m)}{|\mathbf{w}(\delta(m))| + 1},
\end{equation}
where the denominator is equal to the group size. The balance, $B_{cs}(m)$, is then computed as in (\ref{eqn:balance}), with $a_{cs}$ substituted and $a_s$ determined from data.


\begin{figure}[h!]
\centering
\includegraphics[width=\textwidth, angle=270]{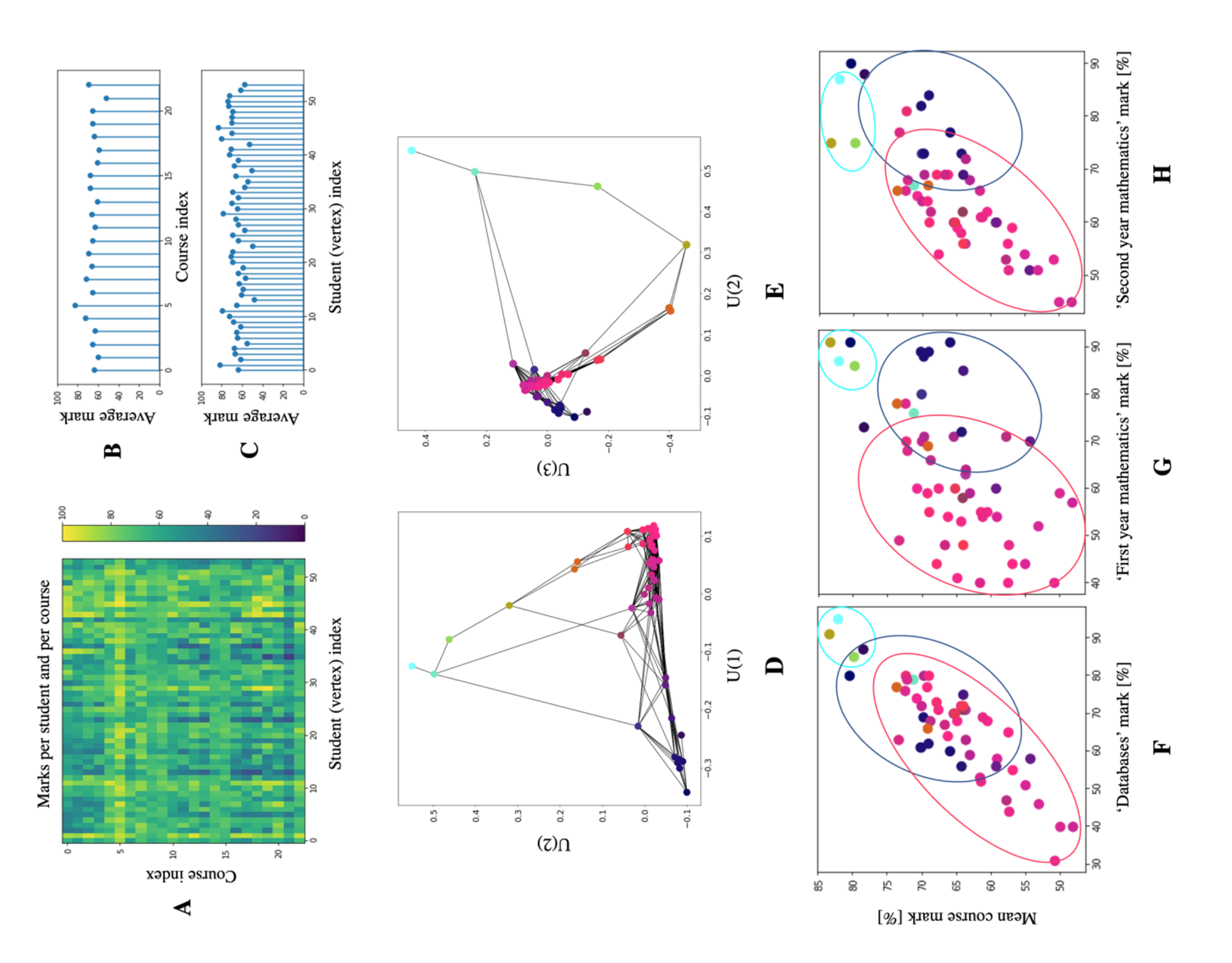}
\caption{The course marks of $N=54$ EIE students in $L=23$ courses viewed as A) marks per student and per course, B) average mark per course, or C) average mark per student. Observe that average marks cannot be used to determine student affinities. To determine student affinities, a graph is constructed with each student represented as a vertex and weighted edges encoding the similarity of course marks between pairs of students. The Laplacian eigenmap for this graph is found, where the dimensionality of the vector describing each student has been reduced from $L = 23$ to $M=3$. D) The eigenmap generated using the first two elements of the spectral vector, $U(1)$ and $U(2)$. E) The eigenmap produced using the second two elements of the spectral vector, $U(2)$ and $U(3)$. To interpret the eigenmap, the mean course mark of each student is plotted against their mark in each course. F) The mark in `databases' course plotted against the mean mark. `Databases' was chosen as a representative example for all other courses excluding mathematics. G) and H) The mark in `first year mathematics' and `second year mathematics' courses plotted respectively against the mean mark. It is observed that students cluster into three affinities: below average in mathematics (pink ellipse), above average in mathematics (dark blue ellipse), and consistent high achievers (teal ellipse). Ellipses are drawn by eye for illustrative purposes. Vertices are coloured by converting the $M=3$ dimensional spectral vector to the RGB triplet.}
\label{fig:eigenmap_process}
\end{figure}

\begin{figure}[t]
\centering
\includegraphics[width=0.7\textwidth]{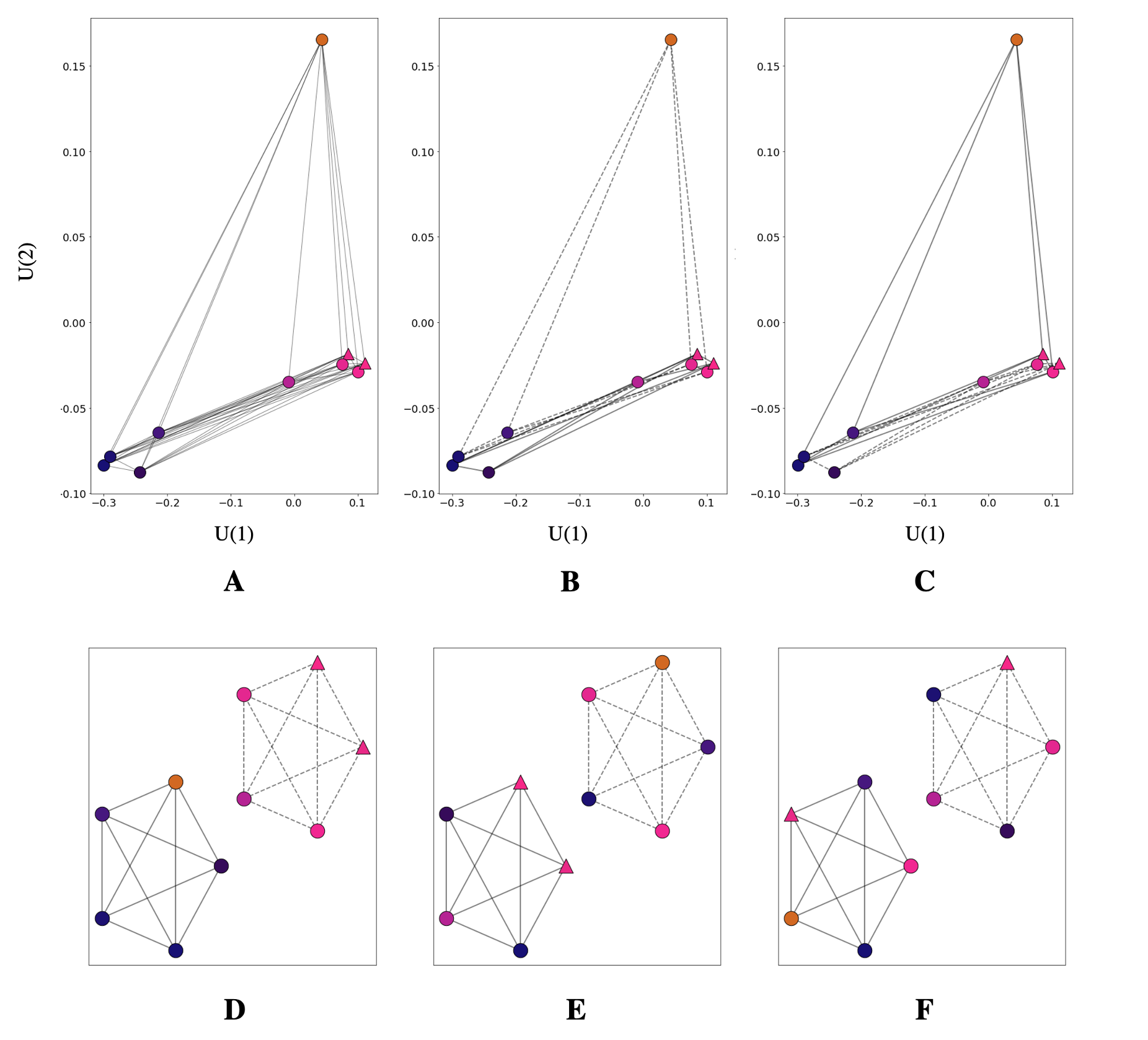}
\caption{Visualisation of student groups in the Laplacian eigenmap (top row) and an arbitrary space (bottom row). A) 2D Laplacian eigenmap with subset of 10 students plotted as vertices with edges defined using (\ref{eqn:graph_construction}). B) and E) Graph partition for skill-diverse group formation found by maximising the objective in (\ref{eqn:optimise}), subject to a group size constraint $F_L = F_U = 5$. Observe in B) and E) that skill sets (colours) are diversified within groups. C) and F) Graph partition for fair and skill-diverse group formation found by maximising the objective in (\ref{eqn:optimise}), subject to a group size constraint $F_L = F_U = 5$ and a balance constraint $B_{cs} = 1$. Observe in C) and F) that skill sets are diversified within groups and students with a sensitive attribute are separated into different groups. D) Graph partition found by minimising the objective in (\ref{eqn:optimise}), subject to a group size constraint $F_L = F_U = 5$. Triangular vertices represent students with a sensitive attribute. Solid and dashed edges connect vertices in different groups. Vertices are coloured by converting the $M=3$ dimensional spectral vector to the RGB triplet.}
\label{fig:diversified_fair_clustering}
\end{figure}

\begin{figure}[t]
\centering
\includegraphics[width=0.7\textwidth]{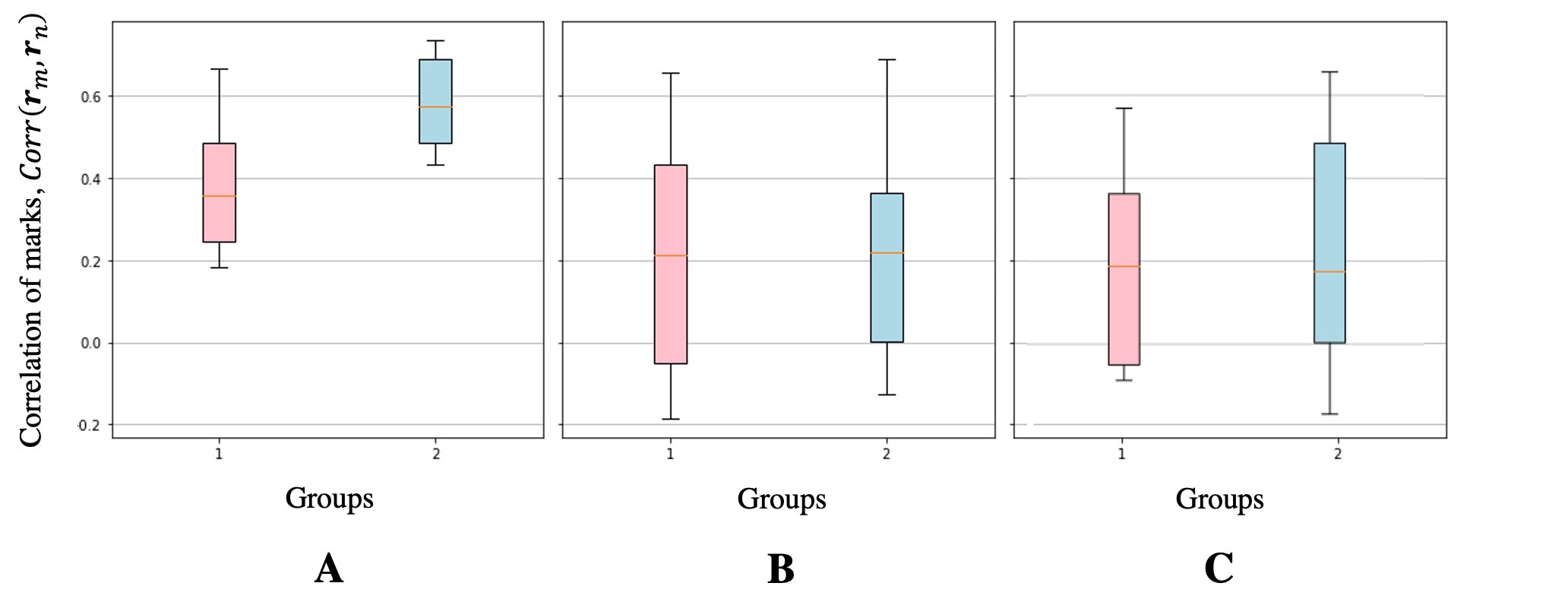}
\caption{Boxplots showing the correlation between course marks, $Corr(\mathbf{r}_m, \mathbf{r}_n)$, for all pairs of students $m$ and $n$ within each group. A) Graph partition found by minimising the objective in (\ref{eqn:optimise}), subject to a group size constraint $F_L = F_U = 5$. B) Graph partition for skill-diverse group formation found by maximising the objective in (\ref{eqn:optimise}), subject to a group size constraint $F_L = F_U = 5$. C) Graph partition for fair and skill-diverse group formation found by maximising the objective in (\ref{eqn:optimise}), subject to a group size constraint $F_L = F_U = 5$ and a balance constraint $B_{cs} = 1$. }
\label{fig:boxplots}
\end{figure}

\section{Results}
\label{sec:results}
\subsection{Dimensionality reduction using Laplacian eigenmap}
Figure \ref{fig:eigenmap_process}A shows the dataset in tabular form, where the columns contain the marks for every student. The average marks per course and per student are shown in Figures \ref{fig:eigenmap_process}B and \ref{fig:eigenmap_process}C. Observe that average marks cannot be used to determine student affinities. 

A graph is constructed from the EIE dataset of course marks for $N=54$ students and $L=23$ courses according to (\ref{eqn:graph_construction}). The dimensionality of student course marks is reduced from $L=23$ to $M=3$ using the Laplacian eigenmap, where the basis of the Laplacian eigenmap is formed using the spectral vector in (\ref{eqn:spectral_vector}). The Laplacian eigenmap for EIE students is shown in Figures \ref{fig:eigenmap_process}D and \ref{fig:eigenmap_process}E, where different clusters of students are visible. Approximately three clusters were identified and are found to belong to the three affinities shown in Figures \ref{fig:eigenmap_process}F-H. These are: below average in mathematics, above average in mathematics, and consistent high achievers.

\subsection{Fair and skill-diverse student group formation}
For computational ease, ten students were chosen at random from the EIE dataset in order to test the proposed algorithm. The locations of these ten students in the Laplacian eigenmap are shown in Figure \ref{fig:diversified_fair_clustering}A. Two students with the same affinity, below average in mathematics, were assigned a synthetic sensitive attribute as shown by triangular vertices in Figure \ref{fig:diversified_fair_clustering}A. The ratio of the sensitive attribute in this test dataset was $a_s=0.2$.

Our proposed constrained integer programming optimisation procedure for fair and skill-diverse group formation was run by maximising the objective in (\ref{eqn:optimise}). Group sizes were constrained as $F_L = F_U = 5$, and balance was constrained with the lower bound $B_{cs} = 1$, i.e. equal ratio of sensitive attributes in all groups and population. \textsc{OR-Tools} \cite{ortools} was used to conduct the constrained integer programming optimisation. The results of this optimisation are shown in the Laplacian eigenmap in Figure \ref{fig:diversified_fair_clustering}C and in the arbitrary space in Figure \ref{fig:diversified_fair_clustering}F, where solid and dashed edges connect vertices in the two separate groups formed. From the Figures \ref{fig:diversified_fair_clustering}C and \ref{fig:diversified_fair_clustering}F, observe that student groups were formed by connecting students across the Laplacian eigenmap and that balance of the sensitive attribute was enforced by allocating these students to different groups, i.e. $a_{1s} = a_{2s} = a_s = 0.2$.

To illustrate the effectiveness of the proposed algorithm, we compared the results to two alternative optimisation procedures: 1) minimising the objective in (\ref{eqn:optimise}) (minimal diversity) and 2) maximising the objective in (\ref{eqn:optimise}) without a constraint on balance (maximal diversity and unfair). The constraint on group sizes remains the same, $F_L = F_U = 5$. Optimising for minimal diversity forms the group of students shown in Figure \ref{fig:diversified_fair_clustering}D. Observe that students with the same affinity were allocated to the same group, which is not desirable for work in small groups. Optimising for maximal diversity without a constraint on balance forms the groups in the Laplacian eigenmap in Figure \ref{fig:diversified_fair_clustering}B which are also shown in the arbitrary space in Figure \ref{fig:diversified_fair_clustering}E. In this case, it is obvious that the sensitive attribute has not been taken into account, as it appears with in-group ratios $a_{1s} = 0.4$ and $a_{2s} = 0$ compared to the population ratio of $a_s = 0.2$, leading to unfair group formation. The amount of diversification was quantified by looking at the distribution of $Corr(\mathbf{r}_m, \mathbf{r}_n)$ within each group. This is visualised as boxplots for the three optimisation procedures tested. Minimal diversity is shown in Figure \ref{fig:boxplots}A, maximal diversity without balance constraint in Figure \ref{fig:boxplots}B and maximal diversity with balance constraint in Figure \ref{fig:boxplots}C. When both diversity was maximised and balance constrained, groups 1 and 2 in Figure \ref{fig:boxplots}C had a low median correlation of marks with the values of 0.19 and 0.17, respectively, whilst satisfying the fairness constraint.



\section{Discussion and conclusion}
We have proposed an unsupervised algorithm for fair and skill-diverse student group formation. Student skill sets have been determined from course marks using dimensionality reduction via the Laplacian eigenmap. Fair and skill-diverse student group formation has been formulated as a constrained graph partitioning problem that was solved using integer programming. The in-group distance between students in the Laplacian eigenmap has been maximised, and the group sizes and `balance' of a sensitive attribute have been constrained with upper and lower bounds. The effectiveness of the proposed algorithm in promoting skill diversity and fairness has been demonstrated on a dataset of student course marks from Imperial College London. Our algorithm has been deployed this academic year to form second year tutorial groups in the Electronic and Electrical Engineering department at Imperial College London. Feedback from students and academics will be collected at the end of term and detailed in future work.


\section*{Acknowledgments}
Alexander Jenkins is supported by the UKRI CDT in AI for Healthcare \url{http://ai4health.io} (Grant No. P/S023283/1).

\printbibliography

\end{document}